\PassOptionsToPackage{table,xcdraw}{xcolor}
\documentclass[11pt]{article}

\usepackage[preprint]{acl}

\usepackage{times}
\usepackage{latexsym}
\usepackage[dvipsnames]{xcolor} 
\usepackage[most]{tcolorbox}
\usepackage{longtable}
\usepackage{tabularx}
\usepackage{enumitem}
\usepackage{booktabs} 
\renewcommand{\arraystretch}{1.3}

\usepackage{tabularx} 
\usepackage{tablefootnote}
\usepackage{booktabs}
\usepackage{array}
\usepackage{ragged2e}
\usepackage{hyperref}
\newcolumntype{Y}{>{\RaggedRight\arraybackslash}X}
\newcolumntype{P}{>{\RaggedRight\arraybackslash}p{3.6cm}}

\usepackage[T1]{fontenc}

\usepackage[utf8]{inputenc}

\usepackage{microtype}

\usepackage{inconsolata}

\usepackage{graphicx}
\usepackage{tikz}
\usetikzlibrary{arrows.meta}
\usepackage{float}     

\definecolor{darkblue}{rgb}{0, 0, 0.5}
\hypersetup{colorlinks=true, citecolor=darkblue, linkcolor=darkblue, urlcolor=darkblue}

\usepackage{fdsymbol} 

\newcommand{\Sref}[1]{\S\ref{#1}}

\title{Stop Taking Tokenizers for Granted: They Are Core Design Decisions in Large Language Models}

\author{%
Sawsan Alqahtani$^{\dagger \spadesuit }$\thanks{Equal contribution.} \enspace \enspace 
Mir Tafseer Nayeem$^{\clubsuit}$\footnotemark[1] \\ 
\bf Md Tahmid Rahman Laskar$^{\vardiamondsuit \heartsuit}$ \enspace 
Tasnim Mohiuddin$^{\diamondsuit}$ \enspace 
M Saiful Bari$^{\varheartsuit}$\\
$^\dagger$Princess Nourah Bint Abdulrahman University \enspace
$^\spadesuit$Saudi Data \& AI Authority (SDAIA) \enspace \\
$^\clubsuit$University of Alberta  \enspace
$^\vardiamondsuit$Dialpad \enspace
$^\heartsuit$York University \enspace \\
$^\diamondsuit$Qatar Computing Research Institute \enspace
$^\varheartsuit$Amazon AGI\\
\texttt{saalqhtani@pnu.edu.sa, mnayeem@ualberta.ca}
}

\begin{document}
\maketitle

\begin{abstract}
Tokenization underlies every large language model, yet it remains an under-theorized and inconsistently designed component. Common subword approaches such as Byte Pair Encoding (BPE) offer scalability but often misalign with linguistic structure, amplify bias, and waste capacity across languages and domains. This paper reframes tokenization as a core modeling decision rather than a preprocessing step. We argue for a context-aware framework that integrates tokenizer and model co-design, guided by linguistic, domain, and deployment considerations. Standardized evaluation and transparent reporting are essential to make tokenization choices accountable and comparable. Treating tokenization as a core design problem, not a technical afterthought, can yield language technologies that are fairer, more efficient, and more adaptable.
\end{abstract}


\section{Introduction}
\label{sec:introduction}

Large Language Models (LLMs) have achieved remarkable success across tasks ranging from fluent text generation to advanced reasoning that now pushes the boundaries of agentic behavior~\citep{laskar2023systematic,openai2024gpt4technicalreport,grattafiori2024llama3,geminiteam2025geminifamilyhighlycapable,anthropic2025claude,deepseekai2025deepseekr1incentivizingreasoningcapability,langford2025amazon}. Yet, amid the focus on architectural innovations, training methods, and alignment techniques, a fundamental component of LLM development has been overlooked: \emph{tokenization}, the process of segmenting raw text into discrete units that serve as the interface between text and learned representations. This choice shapes a model’s representation, efficiency, and generalization across tasks, languages, and domains ~\citep{ali-etal-2024-tokenizer,goldman2024unpacking,bommarito2025kl3mtokenizersfamilydomainspecific,altintacs2025toksuite}.

The NLP community continues to rely on subword tokenization methods, particularly Byte Pair Encoding (BPE)~\citep{10.5555/177910.177914,sennrich-etal-2016-neural}, as the de facto standard for managing vocabulary and rare words. This reliance has produced uniform designs that mask trade-offs and ignore task-, language- or domain-specific needs. This fragmentation disrupts linguistic coherence\footnote{See Appendix~\ref{illus:token_fragmentation}} and undermines model performance on compositional tasks such as code generation, multi-step reasoning, and multi-hop question answering~\citep{bostrom2020byte,ahia-etal-2023-languages,petrov2023language,dewangan-etal-2025-every,Ni-Liang-2025}. 

These limitations have prompted a growing body of research to reexamine traditional tokenization practices, questioning long-held assumptions about the necessity of explicit segmentation and the impact of poorly aligned boundaries on model performance~\citep{zouhar-etal-2023-tokenization,goldman2024unpacking,schmidt-etal-2024-tokenization}. Current practices also face methodological shortcomings: obscure design processes and poor documentation hinder reproducibility and adaptation to new domains and languages~\citep{bostrom2020byte,dagan2024getting}. These issues are compounded by the absence of standardized evaluation metrics and diagnostic tools, making it difficult to systematically assess tokenizer effectiveness~\citep{beinborn-pinter-2023-analyzing,ali-etal-2024-tokenizer}. Moreover, treating tokenization as a static, detached process prevents co-adaptation between segmentation and representation learning, limiting potential gains in efficiency and linguistic fidelity~\citep{land-bartolo-2024-fishing,zheng2024enhancing}. Together, these gaps highlight that tokenization remains an open and under-theorized area in LLM development.

In practice, developers face consequential design decisions, including choosing token granularity, defining merge operations, and curating training corpora. As a result, many pipelines default to reusing pretrained tokenizers, a practice that often misaligns with the linguistic characteristics or practical constraints of the target setting~\citep{
chen2023meditron70b,dagan2024getting,nayeem-rafiei-2024-kidlm,seo2025doeslanguagespecifictokenizeraffect,bommarito2025kl3mtokenizersfamilydomainspecific}. This mismatch reinforces inefficiencies and biases, highlighting the need for explicit, principled, context-aware decision-making around tokenizer auditing, reuse, adaptation, or retraining, grounded in systematic evaluation.

\subsection{Our Perspective on Tokenization}

\textbf{\textit{In this paper, we argue that tokenization, despite its maturity, continues to be treated as an afterthought, implemented with insufficient scrutiny, contextual awareness, and methodological rigor.}} While subword tokenization was originally designed to balance the trade-offs between word- and character-level approaches, its 
reuse should not exempt it from renewed examination. We advocate for a shift from ad-hoc practices to a principled, context-aware design and evaluation process, aligned with model architecture and training objectives, elevating tokenization to an essential component of model development. We emphasize that this perspective neither assumes nor prioritizes retraining LLMs from scratch. Such retraining is often infeasible for many practitioners and language communities, particularly in low-resource settings, and even when feasible, may be unnecessary or suboptimal relative to principled reuse or adaptation.

Our position is supported by \citet{dewangan-etal-2025-every}, who show that even with fixed vocabulary sizes, optimizing the segmentation strategy alone can yield substantial gains. Yet, despite widespread reliance on standard tokenizers, the field 
lacks a unified and theoretically grounded approach to its selection, adaptation, and evaluation.
To bridge this gap, we advocate a principled framework that integrates tokenizer–model co-design, standardized evaluation, and transparent documentation. 

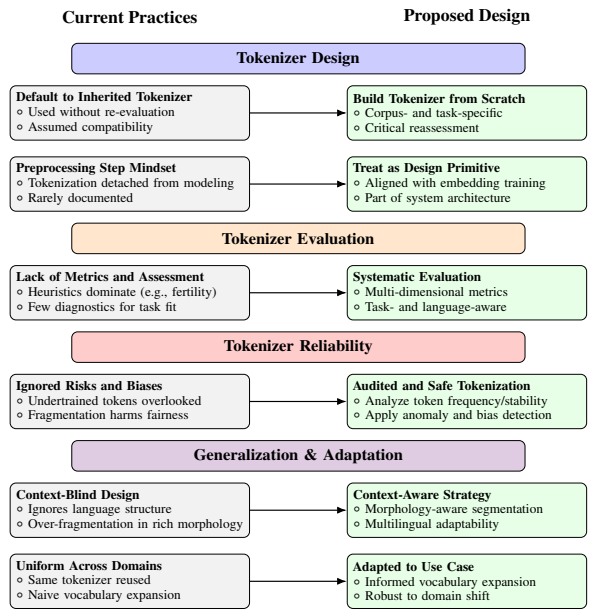
\begin{figure}[!t]
\centering
\resizebox{\linewidth}{!}{
\begin{tikzpicture}[
    font=\small,
    box/.style={rectangle, draw=black, rounded corners,
                text width=5.4cm, minimum height=1.3cm, align=left, inner sep=4pt},
    curr/.style={box, fill=gray!10},
    prop/.style={box, fill=green!10},
    arrow/.style={-Latex, thick},
    label/.style={rectangle, draw=black, rounded corners, fill=#1!20,
                  text width=10.5cm, minimum height=0.73cm, align=center,
                  font=\bfseries\normalsize}
]

\node[font=\bfseries\large] at (-4.0, 14.3) {Current Practices};
\node[font=\bfseries\large] at ( 4.0, 14.3) {Proposed Design};

\node[label=blue] at (0,13.3) {Tokenizer Design};

\node[curr] (c1) at (-4.0,12.0) {\textbf{Default to Inherited Tokenizer}\\
$\circ$ Used without re-evaluation\\
$\circ$ Assumed compatibility};
\node[prop] (p1) at (4.0,12.0) {\textbf{Build Tokenizer from Scratch}\\
$\circ$ Corpus- and task-specific\\
$\circ$ Critical reassessment};
\draw[arrow] (c1.east) -- (p1.west);

\node[curr] (c2) at (-4.0,10.3) {\textbf{Preprocessing Step Mindset}\\
$\circ$ Tokenization detached from modeling\\
$\circ$ Rarely documented};
\node[prop] (p2) at (4.0,10.3) {\textbf{Treat as Design Primitive}\\
$\circ$ Aligned with embedding training\\
$\circ$ Part of system architecture};
\draw[arrow] (c2.east) -- (p2.west);

\node[label=orange] at (0,9.0) {Tokenizer Evaluation};

\node[curr] (c3) at (-4.0,7.7) {\textbf{Lack of Metrics and Assessment}\\
$\circ$ Heuristics dominate (e.g., fertility)\\
$\circ$ Few diagnostics for task fit};
\node[prop] (p3) at (4.0,7.7) {\textbf{Systematic Evaluation}\\
$\circ$ Multi-dimensional metrics\\
$\circ$ Task- and language-aware};
\draw[arrow] (c3.east) -- (p3.west);

\node[label=red] at (0,6.4) {Tokenizer Reliability};

\node[curr] (c6) at (-4.0,5.1) {\textbf{Ignored Risks and Biases}\\
$\circ$ Undertrained tokens overlooked\\
$\circ$ Fragmentation harms fairness};
\node[prop] (p6) at (4.0,5.1) {\textbf{Audited and Safe Tokenization}\\
$\circ$ Analyze token frequency/stability\\
$\circ$ Apply anomaly and bias detection};
\draw[arrow] (c6.east) -- (p6.west);

\node[label=Plum] at (0,3.8) {Generalization \& Adaptation};

\node[curr] (c4) at (-4.0,2.5) {\textbf{Context-Blind Design}\\
$\circ$ Ignores language structure\\
$\circ$ Over-fragmentation in rich morphology};
\node[prop] (p4) at (4.0,2.5) {\textbf{Context-Aware Strategy}\\
$\circ$ Morphology-aware segmentation\\
$\circ$ Multilingual adaptability};
\draw[arrow] (c4.east) -- (p4.west);

\node[curr] (c5) at (-4.0,0.8) {\textbf{Uniform Across Domains}\\
$\circ$ Same tokenizer reused\\
$\circ$ Naive vocabulary expansion};
\node[prop] (p5) at (4.0,0.8) {\textbf{Adapted to Use Case}\\
$\circ$ Informed vocabulary expansion\\
$\circ$ Robust to domain shift};
\draw[arrow] (c5.east) -- (p5.west);

\end{tikzpicture}
}
\caption{{Thematic comparison of current practices and proposed design principles for tokenizer development. 
\vspace{-2.0mm}
}
}
\label{fig:current-proposed}
\end{figure}

As illustrated in Figure~\ref{fig:current-proposed}, such a shift replaces 
rigid practices with a more adaptive and equitable foundation for language technology. Our paper advances this agenda through three contributions:
\begin{itemize}[nosep, topsep=0pt]
\item Identifying the hidden design choices in standard tokenization: vocabulary construction, granularity, corpus selection, and reuse.
\item Proposing a structured tokenizer-design framework that accounts for linguistic diversity, domain specificity, and task complexity.
\item Defining evaluation dimensions 
for fairness, representational alignment, and coverage.
\end{itemize}

By challenging the view of tokenization as an afterthought, we reposition it as a foundational area of empirical inquiry, essential for building language models that are not only efficient but also adaptable and equitable across domains and languages.

\subsection{Alternative Perspectives on Tokenization}
\label{sec:alt_tokenization_views}
Despite growing skepticism toward the \emph{one-size-fits-all} approach, several views continue to shape current tokenization practices~\citep{schmidt-etal-2024-tokenization, reddy2025enough, wegmann-etal-2025-tokenization}. 
We summarize and respond to five common perspectives, outlined in Table \ref{tab:views}.

\begin{table*}[!tb]
\scriptsize
\centering
\setlength{\tabcolsep}{1pt}
\renewcommand{\arraystretch}{1.2} 
\resizebox{\linewidth}{!}  
{
\begin{tabularx}{\textwidth}{>{\raggedright\arraybackslash}p{0.14\textwidth} >{\raggedright\arraybackslash}X}
\toprule
\multicolumn{2}{l}{\textbf{View 1 — Subword Tokenization Is Sufficient and Scalable}}\\
\midrule
\textbf{Claim} &
Subword methods are considered a practical default, offering scalability, compression, and stable training. Their success in general-purpose models has led to the assumption that changing the tokenizer has a minimal impact on LLMs \citep{gutierrez2023biomedical, schmidt-etal-2024-tokenization}.  \\ \hline
\textbf{Response} &
This view downplays the importance of tokenization. In morphologically rich, domain-specific, or code-switched settings, subword tokenization can fragment meaningful units and degrade performance. Yet, tokenizers are often reused with minimal justification, hindering reproducibility. While prior work (e.g., \citealt{schmidt-etal-2024-tokenization}) reports minor performance differences across tokenizers, \citet{schmidt2025boundlessbytepairencoding} suggest this stems from high token overlap, where over 90\% of common words are represented as single tokens, limiting variation to a small input fraction and diminishing any consistent advantage. Moreover, even large models suffer from inefficiencies due to poor tokenization, such as inflated sequence lengths.\footnotemark \\
\midrule
\multicolumn{2}{l}{\textbf{View 2 — Current Evaluation Metrics Accurately Reflect Tokenizer Quality}}\\
\midrule
\textbf{Claim} &
In the context of LLMs, tokenization is often deprioritized because existing evaluations suggest minimal performance differences across tokenizers. It is assumed that model size and contextual depth compensate for poor segmentation, making tokenizer optimization unnecessary. \\ \hline
\textbf{Response} &
This view stems from shallow evaluation practices that fail to capture tokenizer quality. Metrics like vocabulary size, fertility, and compression rate are easy to compute but poorly reflect linguistic alignment or downstream utility (\S\ref{sec:evaluation}). They often obscure issues such as over-fragmentation and poor generalization, especially in multilingual or domain-specific contexts, where similar fertility scores can mask divergent morphological behavior. \\
\midrule
\multicolumn{2}{l}{\textbf{View 3 — Byte-Level Tokenization Removes the Need for Explicit Design}}\\
\midrule
\textbf{Claim} &
This argument has led to claim that explicit tokenizer design is becoming obsolete \citep{feher-etal-2025-retrofitting}. These methods promise universality, avoiding the need for language-specific preprocessing, and simplify the training pipeline by operating on raw text bytes. \\ \hline
\textbf{Response} &
While appealing in their simplicity and universality, byte-level approaches do not eliminate tokenizer design so much as relocate it. By operating over raw bytes, they shift the burden of discovering linguistic, morphological, and semantic structure entirely onto the model. This shift introduces new trade-offs: substantially longer sequences, increased training and inference overhead under standard attention, and greater reliance on model capacity and inductive bias to recover structure \citep{moon2025bitlevelbpebyteboundary}. Although emerging architectures such as linear attention and patch-based processing may mitigate some of these costs, current evidence suggests that sequence length and computational overhead remain significant concerns in practical deployments \cite{pagnoni2024byte}. \\

\midrule
\multicolumn{2}{l}{\textbf{View 4 — Token-Free or Continuous Representations Will Replace Tokenization}}\\
\midrule
\textbf{Claim} &
Emerging ``token-free’’ paradigms argue that tokenization is an artificial bottleneck inherited from discrete symbolic modeling \citep{pagnoni2024byte,hwang2025dynamic,neitemeier2025hierarchical}. Such models aspire to bypass tokenization altogether, potentially enabling end-to-end differentiable pipelines and more fine-grained representational learning. \\ \hline
\textbf{Response} &
While token-free modeling is a promising frontier, it is far from a resolved alternative. Continuous-input systems face challenges in scalability, interpretability, and transferability, particularly when training at web-scale. Moreover, the absence of discrete boundaries complicates linguistic analysis and data processing pipelines, which still rely heavily on symbolic granularity. \\
\midrule
\multicolumn{2}{l}{\textbf{View 5 — Tokenizer–Model Joint Optimization Enables True Co-Design}}\\
\midrule
\textbf{Claim} &
Recent work argues that genuine tokenizer–model co-design demands architectural coupling, treating the tokenizer as a learnable module co-optimized with model parameters \cite{hiraoka2021joint}. In this view, token boundaries and vocabulary adapt dynamically to representational demands, aligning linguistic structure with learned embeddings to improve efficiency and adaptability. \\ \hline
\textbf{Response} &
While conceptually appealing, integrating tokenization into the training loop poses challenges of stability, efficiency, and interpretability. Joint optimization risks entangling representation learning with unstable segmentation dynamics. \\
\bottomrule
\end{tabularx}
}
\caption{Five alternative perspectives on tokenizer design and our responses to each.
\vspace{-3mm}
}
\label{tab:views}
\end{table*}
\footnotetext{See Appendix \ref{appendix:evolution_tokenizer} for examples from leading LLMs.}

\section{Tokenization and Alternatives}


\subsection{Subword Tokenization Methods}
\label{sec:subword_tokenization_methods}

Subword tokenization addresses the drawbacks of word- and character-level models. Word-level approaches suffer from large vocabularies and out-of-vocabulary issues, while character-level models create long sequences that raise computational costs and obscure linguistic structure \citep{beinborn-pinter-2023-analyzing}. Subword methods (e.g., BPE, WordPiece, Unigram) balance these extremes by segmenting text into frequent, data-driven units, enabling efficient handling of rare words and manageable sequence lengths \citep{wolleb-etal-2023-assessing}.

These methods differ in segmentation strategy. BPE and WordPiece are merge-based: BPE joins frequent adjacent pairs by frequency, while WordPiece optimizes a likelihood-based objective tied to language modeling \citep{schmidt-etal-2024-tokenization}. Both yield deterministic outputs. In contrast, Unigram takes a probabilistic approach, pruning subwords with low data likelihood and allowing multiple valid segmentations. This often makes Unigram more flexible and linguistically aligned, especially for multilingual or morphologically rich languages~\citep{kudo-2018-subword, bostrom2020byte}.

BPE offers simplicity and speed, WordPiece emphasizes likelihood-based merge decisions informed by local token context, and Unigram provides cross-lingual adaptability at higher computational cost. These methods underpin models like GPT \citep{brown2020languagegpt3}, BERT \citep{devlin-etal-2019-bert}, and LLaMA \citep{touvron2023llama,touvron2023llama2}. Yet their reliance on frequency statistics exposes key limits~\citep{meyer-buys-2022-subword,liu-etal-2024-gold,chai-etal-2024-tokenization} frequent forms dominate, while rare words, low-resource languages, and domain-specific terms are poorly captured. Token boundaries often mirror statistical artifacts rather than linguistic structure, impairing generalization in typologically complex languages~(see Appendix~\ref{illus:token_fragmentation}). Despite scalability, these methods often underperform in compression efficiency and cross-lingual representational alignment \citep{goldman2024unpacking}.

\vspace{-2mm}
\subsection{Emerging Alternatives for the De Facto Tokenization Paradigm}
\label{alternative_approaches}
Tokenization based on statistical frequency over large corpora has long dominated NLP workflows, but it struggles with rare languages, domain-specific terms, and structural linguistic nuances~\citep{meyer-buys-2022-subword,liu-etal-2024-gold,chai-etal-2024-tokenization}. In response, the research community has proposed several alternatives that challenge the subword-centric status quo. Without endorsing a particular solution, we review these approaches to underscore a broader shift toward more adaptive, linguistically informed strategies and to highlight open questions that remain.

\paragraph{Learning Compositional Structure Explicitly:} Traditional tokenization ignores linguistic composition, treating subwords as independent units \citep{bostrom2020byte}. Recent methods attempt to model internal structure more directly through sparse embeddings \citep{deiseroth2024t} or morpheme-aware GRU frameworks \citep{singh2023subwords}, but these approaches add training complexity and remain underexplored in large-scale LLMs.



\paragraph{From Subwords to Raw Bytes:} Subword tokenizers struggle with unseen characters and typographic variation~\citep{xue-etal-2022-byt5}. Byte-level methods address this by operating directly on raw input, avoiding fixed vocabularies. The Byte-Level Transformer (BLT)~\citep{pagnoni2024byte} employs entropy-based patches to reduce inference cost and remove vocabulary constraints. While promising, these methods can increase sequence length and computational overhead.


\paragraph{Modeling Language at  Semantic Level:}
Recent work explores higher-level abstractions by representing multi-word expressions or concepts as single units~\citep{liu2025superbpe,schmidt2025boundlessbytepairencoding}. \citet{liu2024generation} propose a dynamic phrase encoder with variable-length phrases to enhance generalization, while \citet{lcm2024} introduce sentence-level embeddings that bypass token-level processing and view language as a hierarchy of meaning~\citep{shao2025beyond}, benefiting multilingual tasks but facing data sparsity challenges.



\section{Rethinking Tokenizer Design: From Pitfalls to Practice}

Building on prior findings that highlight tokenization’s influence on model behavior~\citep{bostrom2020byte,ali-etal-2024-tokenizer}, we identify four persistent pitfalls in current practice.


\paragraph{Design Integration:}
Tokenizer design should be integrated into model training, not inherited as a fixed preprocessing artifact. Reliance on legacy vocabularies, often optimized for different data or languages, limits control and reproducibility. Effective design begins with a core choice: train a tokenizer from scratch
, or reuse an existing one for efficiency and interoperability (\Sref{sec:training-tokenizer}, \Sref{sec:adapting-tokenizer}).

\paragraph{Adaptation and Generalization:}
Tokenizers are frequently reused across domains without adaptation, resulting in representational gaps and inflated sequence lengths. As discussed in \Sref{sec:adapting-tokenizer}, domain- and language-aware tokenization is essential for fair and robust generalization across linguistic and structural variation.

\paragraph{Evaluation Practices, Reliability, and Bias:}
Tokenizer evaluation often relies on surface metrics such as 
fertility, which correlate weakly with performance~\citep{zouhar-etal-2023-tokenization, ali-etal-2024-tokenizer}. As detailed in \Sref{sec:evaluation}, systematic assessment across linguistic fit, task alignment, and cross-lingual robustness is needed. Current workflows overlook token fragmentation, undertrained embeddings, and biases against morphologically rich or low-resource languages. Responsible tokenization requires auditable workflows, including anomaly detection and vocabulary stability checks (\Sref{sec:responsible-tokenization}).

\vspace{-1mm}

\subsection{A Structured Iterative Process}
\label{sec:iterative}
\vspace{-1mm}
An effective tokenizer should meet empirically grounded desiderata, ensuring that vocabulary units occur frequently enough during training to support robust learning and practical utility:
\vspace{-1mm}
\begin{tcolorbox}[
colback=gray!7, colframe=black!11,
title=Tokenizer Design Desiderata,
fonttitle=\bfseries\small, coltitle=black,
]
\small
\textbf{•} ~\textbf{Linguistically Faithful:} mirrors  structure, morphology, syntax, and code-switching patterns.

\textbf{•}~\textbf{Usage-Aligned:} matched to intended language coverage, domains, and 
downstream usage profiles.

\textbf{•}~\textbf{Deployment-Ready:} efficient under latency and memory limits.

\textbf{•}~\textbf{Architecturally Compatible:} fits model architecture and embeddings.
\end{tcolorbox}
\vspace{-1mm}

To move beyond ad hoc design, we propose a structured, iterative co-design process in which tokenization is periodically re-evaluated and refined alongside model architecture and training objectives. Tokenization is informed indirectly by model behavior, using representational analyses and downstream performance as feedback. The workflow integrates linguistic analysis and empirical evaluation, with each stage refining earlier choices to maintain coherence across linguistic, operational, and deployment constraints. Table~\ref{tab:tokenizer_development} summarizes the core stages and outputs. While this process offers a practical roadmap, success hinges on how language is segmented and represented. The goal is not to over-specialize tokenization to a particular benchmark, but to avoid blindly inheriting a tokenizer misaligned with the target language or domain~\citep{nayeem2025STRR}.

\begin{table*}[t]
\centering
\scriptsize
\resizebox{16cm}{!}  
{
\begin{tabular}
{p{1.9cm}p{7.1cm}p{2.3cm}p{3.5cm}}
\toprule
\textbf{Stage} & \textbf{Key Activities} & \textbf{Expected Output} & \textbf{Illustrative Example} \\
\midrule

\textbf{Initial Assessment} &
\textbf{(i)} Define the intended LLM purpose (e.g., general or domain-specific). \newline
\textbf{(ii)} Identify target languages, dialects, and cultural contexts. \newline
\textbf{(iii)} Specify required capabilities (e.g., reasoning). \newline
\textbf{(iv)} Establish model constraints (architecture, memory, latency). 
& 
A clear set of design objectives and operational constraints to guide tokenizer and vocabulary development.
&
Designing a biomedical-domain LLM focusing on English and Spanish medical notes, with strict latency constraints for on-device clinical use.
\\
\hline

\textbf{Reuse Decisions} &
\textbf{(i)} Identify existing tokenizers with appropriate licenses. \newline
\textbf{(ii)} Evaluate against linguistic fidelity, task alignment, and compatibility. \newline
\textbf{(iii)} Run diagnostics (e.g., vocabulary overlap, token length distribution). \newline
\textbf{(iv)} Decide to reuse 
without changes, adapt, or create a new tokenizer.
&
A diagnostic report and an evidence-based decision to reuse, adapt, or develop a new tokenizer.
&
Comparing GPT-2 and LLaMA tokenizers for biomedical abstracts, identifying 
the vocabulary overlap, and deciding to train a new tokenizer for domain-specific terminology.
\\
\hline

\textbf{Corpus Curation} &
\textbf{(i)} Build corpora aligned with the intended use case. \newline
\textbf{(ii)} Establish diagnostic criteria (coverage, diversity, task relevance).
&
A representative, task-aligned corpus.
&
Curating a multilingual biomedical corpus of clinical trial reports, medical guidelines, and radiology notes.
\\
\hline

\textbf{Pretokenization \& Processing} &
\textbf{(i)} Apply pretokenization rules and preprocessing steps. \newline
\textbf{(ii)} Normalize encoding, clean text, set boundary rules. \newline 
\textbf{(iii)} Handle code-switching, numbers, and mixed inputs. \newline
\textbf{(iv)} Align with application requirements.
&
A documented, reproducible set of pretokenization rules applied consistently.
&
Normalizing biomedical text by handling hyphenated terms (e.g., “HER-2/neu”), gene names, and units of measurement.
\\
\hline

\textbf{Tokenizer Training} &
\textbf{(i)} Select a segmentation strategy. \newline
\textbf{(ii)} Train tokenizer on the curated corpus to produce vocabulary.
&
A representative, task-aligned corpus and a trained tokenizer with an initial vocabulary.
&
Training a 60K-token BPE tokenizer on 25 GB of biomedical literature to capture domain-specific compounds (e.g., “immunohistochemistry”).
\\
\hline

\textbf{Intrinsic \& Extrinsic Evaluation} &
\textbf{(i)} Apply tokenizer to varied datasets. \newline
\textbf{(ii)} Compute metrics (e.g., token length, fragmentation). \newline
\textbf{(iii)} Audit vocabulary and visualize anomalies. \newline
\textbf{(iv)} Evaluate edge cases and small model performance.
&
A comprehensive evaluation report covering tokenizer-level diagnostics and model-based validation.
&
Measuring token length across English and Spanish biomedical texts; identifying over-segmentation in long clinical terms.
\\
\hline

\textbf{Iteration \& Model-Guided Refinement} &
\textbf{(i)} Refine tokenizer using diagnostics. \newline
\textbf{(ii)} Treat tokenizer as co-evolving with the model. \newline
\textbf{(iii)} Re-run evaluations until stable performance.
&
A refined, model-aligned tokenizer with improved efficiency and quality.
&
After observing poor tokenization of biomedical abbreviations, re-merging tokens to preserve critical acronyms like “MRI” and “ICU.”
\\
\hline

\textbf{Final Integration \& Documentation} &
\textbf{(i)} Confirm stability and finalize configuration. \newline
\textbf{(ii)} Document known limitations and design rationale. \newline
\textbf{(iii)} Archive and publish tokenizer for reproducibility.
&
A production-ready tokenizer with comprehensive documentation.
&
Publishing biomedical tokenizer specifications with open access, including examples for drug names and clinical abbreviations.
\\
\bottomrule
\end{tabular}
\vspace{-5mm}
}
\caption{{Core Stages in Tokenizer Development Process, extended with illustrative examples.
}
}
\label{tab:tokenizer_development}
\end{table*}


\subsection{Critical Decisions within the Workflow}
\label{sec:critical-decisions}

Each of the pitfalls identified earlier maps to design decisions that shape tokenizer quality. We unpack these decisions, showing how principled choices throughout tokenizer development can systematically address the deficiencies outlined above. Each subsection concludes with guiding questions, consolidated in Table \ref{tab:guiding_questions} in the Appendix.


\subsubsection{Training Tokenizers from Scratch}
\label{sec:training-tokenizer}

Designing a tokenizer from scratch enables deliberate control over segmentation, corpus, vocabulary size, and preprocessing, aligning it with linguistic, functional, and deployment needs~\citep{dagan2024getting}. In contrast, reusing a pretrained model’s tokenizer (e.g., LLaMA) constrains these choices, often limiting performance and generalization. This section outlines key design factors and open questions for principled tokenizer development.

\paragraph{Choosing the Right Tokenizer Algorithm.} 
Selecting a tokenization algorithm is a crucial early step but often defaults to convenience over task or language needs~\citep{ali-etal-2024-tokenizer}. Unigram better captures morpheme-level structure in agglutinative languages~\citep{bostrom2020byte}, while frequency-based BPE remains dominant. These choices influence representation quality, sequence length, and generalization~\citep{ahia-etal-2023-languages}. Recent work shows that tokenizers tailored to linguistic and operational contexts outperform generic ones, especially in domain-specific and multilingual settings~\citep{bommarito2025kl3mtokenizersfamilydomainspecific}. Evidence also suggests that multilingual tokenizers preserve morphology better in complex languages than in simpler ones~\citep{arnett-bergen-2025-language}, though the benefits of morphologically aligned tokenization remain debated~\citep{arnett2025evaluating}.

\paragraph{Training Data Matters for Tokenizer Effectiveness.}
Training corpus composition critically shapes tokenizer performance~\citep{reddy2025enough}. Reliance on large-scale web crawls introduces representational biases, particularly for low-resource languages and specialized domains, where vocabularies built from unrelated or noisy data degrade performance~\citep{rust-etal-2021-good,goldman2024unpacking}. Yet metrics for corpus quality remain limited. Empirical studies show that compression quality, driven by corpus composition and segmentation strategy, correlates strongly with model utility, especially for rare or domain-specific terms~\citep{goldman2024unpacking}. Evidence also suggests tokenizer quality plateaus beyond roughly 150–180GB of training data, indicating diminishing returns from further scaling~\citep{reddy2025enough}.


\paragraph{Pre-tokenization and Normalization Are Foundational.}
Pre-tokenization and normalization encode linguistic assumptions that shape token consistency, sequence length, and vocabulary fragmentation~\citep{dagan2024getting}. Even minor choices (e.g., Unicode normalization or diacritic removal) can markedly alter segmentation.\footnote{E.g., normalizing “coöperate” to “cooperate” avoids inconsistent splits in OCR or historical texts.} They also guide subword merges via boundary cues: incorporating multiword expressions or local context yields more coherent merges, especially in agglutinative languages, often requiring BPE adjustments~\citep{schmidt-etal-2024-tokenization}. Removing pre-tokenization consistently degrades performance~\citep{dagan2024getting,velayuthan-sarveswaran-2025-egalitarian}, sometimes more than changing vocabulary size or corpus composition~\citep{wegmann-etal-2025-tokenization}, and can create indivisible units that limit gains from larger corpora~\citep{reddy2025enough}.


\paragraph{Vocabulary Size Is a Strategic Design Choice.}
Vocabulary size should be co-optimized with model architecture and training goals. Larger vocabularies reduce fragmentation and sequence length but raise memory and embedding costs~\citep{tao2024scaling}. These trade-offs intensify in multilingual or morphologically rich settings~\citep{ali-etal-2024-tokenizer}. Studies show that scaling vocabulary can improve compression and efficiency, though gains depend on corpus, task, and model capacity~\citep{tao2024scaling,huang2025overtokenizedtransformervocabularygenerally}. Yet, \citet{schmidt2025boundlessbytepairencoding} warns that adding low-frequency tokens may increase redundancy with minimal benefit. Despite this, vocabulary size is often set heuristically~\citep{salesky2020optimizing,gowda-may-2020-finding}. A more principled approach would treat it as a tunable parameter informed by token frequencies, learning dynamics, and compute limits, though key questions remain.


\paragraph{Open Trade-offs.}
Table~\ref{tab:train-from-scratch} summarizes approaches that learn tokenizers directly from raw text instead of reusing legacy vocabularies~\citep{sennrich-etal-2016-neural,kudo-2018-subword,kudo-richardson-2018-sentencepiece,tay2022charformer}. When data and budget are controllable, first-principles design is often preferable~\citep{gowda-may-2020-finding,zouhar-etal-2023-formal}. Vocabulary size should be treated as a computational lever, not a sign of model ``purity'': smaller vocabularies inflate sequence length and attention cost, while larger ones increase parameters and memory with diminishing returns~\citep{gowda-may-2020-finding,zouhar-etal-2023-formal}. Classic subword methods (BPE, WordPiece, Unigram) remain strong baselines for minimizing tokens per sample~\citep{sennrich-etal-2016-neural,kudo-2018-subword,kudo-richardson-2018-sentencepiece}. Morphology-aware or lexically grounded variants are worthwhile for agglutinative or templatic languages~\citep{libovicky-helcl-2024-lexically,chizhov-etal-2024-bpe,asgari2025morphbpe}, while block-level schemes such as Charformer downsampling retain character-level robustness without excessive sequence inflation~\citep{tay2022charformer}.




\subsubsection{Adapting Tokenizers: Reuse Requires Verification and Documentation}
\label{sec:adapting-tokenizer}

Reuse of mature, pretrained tokenizers is often a reasonable and practical default, particularly when compatibility, distillation from existing models, and engineering efficiency are primary concerns. Whether a pretrained tokenizer can serve as-is or requires adaptation depends on its alignment with the target language or domain. When LLMs are extended to new settings, work typically centers on weight fine-tuning, yet performance often degrades because English-optimized tokenizers fail to capture domain terms and morphology in low-resource contexts~\citep{sha-etal-2025-veef}. Vocabulary expansion is a common but ad hoc remedy, with limited understanding of when it helps~\citep{wang-etal-2019-improving,kiulian2024english}, often yielding short-term gains that fail to generalize.

Robust adaptation instead requires treating tokenization as a core design task, guided by systematic protocols for token addition, embedding initialization, and adaptation staging. While reuse improves compatibility and cost, unexamined reuse can introduce long-term drawbacks, including loss of linguistic fidelity, coverage gaps, and fairness issues, particularly when the tokenizer’s original training data or design assumptions diverge from the new setting. Crucially, although training or adapting a tokenizer incurs a one-time engineering cost, inheriting a suboptimal tokenizer imposes a persistent inference-time penalty: inflated sequence lengths increase latency and compute per query over the lifetime of a deployed model, especially in non-English and domain-specific scenarios (see Appendix~\ref{illus:token_fragmentation}).


\paragraph{Base Model Choice: \emph{Tokenizer Compatibility Is Not Guaranteed.}}

Adaptation often starts with choosing a base LLM and reusing its tokenizer by default, yet the suitability of this choice for a new domain or language is rarely explicitly tested or documented. Even when vocabularies are expanded, such as extending the LLaMA 2 tokenizer from 32K to 61K tokens for Arabic~\citep{bari2025allam}, the linguistic rationale behind token boundaries remains unclear. Without evaluating what the original tokenizer captures or omits, expansion risks becoming a workaround.

\paragraph{{Vocabulary Size:} \emph{Incremental Growth Without Clear Guidance.}}
In adaptation, expanding vocabulary means adding new tokens to an existing tokenizer. This can shorten sequences and improve representation for new domains or languages, but also raises issues of compatibility, memory, and convergence~\citep{tejaswi2024exploring}. Unlike tokenizers built from scratch, expansion operates within the constraints of an inherited scheme. Evidence suggests that targeted additions (e.g., 5–10K tokens) yield better returns than aggressive scaling~\citep{liu-etal-2024-gold}, yet consistent methods for selecting, validating, and integrating tokens remain lacking. Without a grounded approach, expansions risk inefficiency and misalignment with the base model’s learned representations. Because expansion preserves the original token inventory, it is often preferred in settings where distillation from an existing model, checkpoint reuse, or interoperability with deployed systems is required.

\paragraph{{Initialization:} \emph{A Crucial but Understudied Factor.}}
How new token embeddings are initialized strongly affects downstream performance. Approaches include random initialization, subtoken averaging, and using external word vectors~\citep{yao-etal-2021-adapt,dobler2023focus}, but comparative evidence remains limited and context-dependent. Stopping criteria such as fertility scores are often borrowed from unrelated tasks and rarely validated across languages. The field needs benchmarks and reporting standards to identify which strategies generalize and under what conditions.

\paragraph{{Embedding Warm-Up:} \emph{A Temporary Fix or Best Practice?}} Embedding warm-up (training new token embeddings in isolation before unfreezing the full model) has emerged as a practical compromise~\citep{zhao2024llamaenglishempiricalstudy}. Some workflows add phased unfreezing of input or output layers~\citep{kim2024efficienteffectivevocabularyexpansion}, but these remain empirical heuristics rather than principled frameworks. Without systematic evaluation, it remains unclear whether such strategies are robust solutions or fragile stopgaps that conceal deeper misalignments.

\paragraph{Open Trade-offs.}
Table~\ref{tab:adapt-reuse} summarizes tokenizer adaptations across languages and domains. Blind reuse of legacy tokenizers often inflates sequence length through over-fragmentation, wasting context and compute. In contrast, retokenization, vocabulary expansion, and merge refinement are offline interventions that typically repay their cost with shorter sequences and better context use~\citep{rust-etal-2021-good,yehezkel-pinter-2023-incorporating,chizhov-etal-2024-bpe}. Large-scale studies show that tokenizer choice materially affects both quality and training cost, and that naïve reuse can increase training steps across languages~\citep{ali-etal-2024-tokenizer,schmidt-etal-2024-tokenization}. Adaptations in Arabic and Southeast Asian settings demonstrate that targeted expansions or dedicated tokenizers close token-length gaps while preserving English performance~\citep{nguyen2023seallms,bari2025allam,team2025fanar}. For unseen scripts, compact in-language tokenizers with aligned embeddings reduce UNKs and avoid costly character fallback~\citep{pfeiffer-etal-2021-unks}. Taken together, these findings suggest that reuse should be treated as a default hypothesis to be tested, rather than an unquestioned assumption.
\subsubsection{Responsible Tokenization: Confronting Risks to Fairness, Stability, and Security}
\label{sec:responsible-tokenization}
Most tokenizer discussions emphasize performance, but robust and responsible design is equally critical. Tokenization influences fairness, stability, and security, shaping language coverage, equity, and susceptibility to attacks. Fairness refers to avoiding unequal segmentation quality, representational accuracy, or computational burden across languages, scripts, or demographic groups, ensuring parity in token lengths and downstream performance. Stability refers to how consistently and reliably a tokenizer segments text across noise, scripts, and domains, while avoiding low-quality or undertrained units. As LLMs enter real-world deployment~\citep{laskar2023building,fu-etal-2024-tiny,nayeem2025opiniorag}, tokenization must be treated as a potential source of bias for transparent design.


\paragraph{{Undertrained Tokens:} \emph{Silent Failures in the Vocabulary.}}
Undertrained tokens, units that occur infrequently in the pretraining corpus, receive too few updates during training. They often stem from rare words, domain-specific terms, or overly fragmented vocabularies. Their effects are subtle but harmful, causing instability and poor generalization~\citep{Yang2024ProblematicTT}. The notorious SolidGoldMagikarp token, which emerged in GPT models due to skewed data distributions, illustrates how poorly trained tokens can trigger erratic behavior~\citep{solidgoldmagikarp}. Yet such tokens are seldom identified. 
Embedding audits, token pruning, vocabulary refinement, and frequency analysis should be standard in tokenizer evaluation~\citep{chizhov-etal-2024-bpe,cognetta2024analysis,bauwens-delobelle-2024-bpe,lian2025scaffold}. 


\paragraph{Anomalous Tokens: \emph{Artifacts That Disrupt Learning.}}
Tokenizer vocabularies often contain anomalous tokens, units introduced by data noise, encoding errors, or preprocessing inconsistencies. These may capture HTML fragments, URL encodings, or control characters that add no linguistic value and instead degrade embedding quality~\citep{jiang2024investigating,chai-etal-2024-tokenization}. Their presence injects noise into training, causing unpredictable behavior in generation and classification. More broadly, such artifacts pose challenges for reproducibility, auditing, and verification across machine learning pipelines, where undocumented preprocessing decisions can silently affect downstream behavior. While this lack of validation is a broader issue in model reproduction and auditing, tokenizers are especially affected because vocabulary artifacts become fixed and propagate throughout training and deployment. Clearer standards for corpus preparation, token quality control, and automated anomaly detection are therefore essential not only for tokenizer quality, but for reproducible, auditable, and trustworthy model development. 

\paragraph{Tokenization Bias: \emph{ Unequal Representation by Design.}}
Tokenization may not treat all languages or identity groups equally. Morphologically rich languages often suffer excessive fragmentation, lengthening sequences and degrading performance~\citep{toraman2023impact}~(see Appendix~\ref{illus:token_fragmentation}). Similarly, identity-related terms from underrepresented regions are frequently over-segmented, impairing tasks like named entity recognition~\citep{ahiamagnet}. These are not random artifacts but outcomes of design choices that allocate vocabulary space unequally across languages and demographics. Responsible tokenizer development must therefore ensure balanced language representation, introduce fairness constraints, and systematically analyze segmentation across diverse groups. Without such efforts, tokenization remains a structural source of bias in downstream applications.

\paragraph{Security and Adversarial Risks: \emph{{A Hidden Attack Surface.}}}
Tokenizer design can introduce overlooked security vulnerabilities. Adversarial token sequences, prompt injection, and homograph attacks exploit weaknesses in tokenization rules and preprocessing~\citep{wang2024tokenization, jang-etal-2025-improbable}. Poor handling of visually similar characters across scripts further enables such exploits. Security should not be an afterthought: tokenization must be included in adversarial testing, with controlled preprocessing, careful treatment of special tokens, and rigorous auditing and versioning. As LLMs enter sensitive domains, securing tokenization is essential for both performance and trust.

\paragraph{Open Trade-offs.}
As shown in Table~\ref{tab:responsible}, tokenizer design is a governance decision, not incidental plumbing. It determines sequence length, training cost, and robustness to noise and cross-language parity~\citep{rust-etal-2021-good,petrov2023language,ahia-etal-2023-languages}. Minor orthographic noise or typos 
can destabilize segmentation~\citep{chai-etal-2024-tokenization}, 
while morphology-aware constraints and merge refinement improve stability and parity with minimal runtime overhead~\citep{chizhov-etal-2024-bpe,asgari2025morphbpe}. 
Recent studies show that targeted adaptations reduce token-length gaps in Southeast Asian and Arabic settings without harming English performance~\citep{nguyen2023seallms,bari2025allam}.



\subsubsection{Evaluation Framework and Gaps}
\label{sec:evaluation}

Despite tokenization’s central role in model performance, evaluation remains limited and fragmented. Most work relies on surface metrics (compression rate, vocabulary size, fertility) that are easy to compute but reveal little about linguistic alignment or generalizable learning~\citep{ali-etal-2024-tokenizer, laskar2024systematic,
uzan-etal-2024-greed, nayeem2025STRR}. This gap is most pronounced in multilingual and domain-specific settings, where similar fertility scores can hide large differences in morphological preservation or rare-construction handling~\citep{bostrom2020byte}. These metrics are also sensitive to corpus composition, language typology, and preprocessing, limiting cross-domain validity~\citep{rust-etal-2021-good}. Without a principled evaluation framework, trade-offs among linguistic fidelity, efficiency, 
and robustness remain poorly understood. Tokenization choices are seldom tied to downstream performance~\citep{dagan2024getting}.

Notably, many of these diagnostics (such as token-per-character statistics by language, fragmentation analysis, anomaly detection, and undertrained-token audits) can be applied directly to existing pretrained models without retraining, making them practical even under tight compute or data constraints. We do not argue that tokenizer evaluation requires training large models to convergence, but instead advocate low-cost intrinsic and model-based probes (e.g., token-per-character parity and 
information-theoretic efficiency~\cite{tsvetkov-kipnis-2024-information}) that correlate with downstream multilingual performance. 

\vspace{-1mm} 

\paragraph{Toward Multi-Dimensional Evaluation.}
Recent work has questioned the validity of common intrinsic metrics. Early studies hypothesized that reducing corpus token counts (CTC) would improve downstream performance~\citep{galle-2019-investigating,goldman2024unpacking}, but later findings show no consistent correlation between CTC, vocabulary size, and task accuracy~\citep{schmidt-etal-2024-tokenization,ali-etal-2024-tokenizer}. In response, researchers have proposed richer measures of token quality. \citet{zouhar-etal-2023-tokenization} introduce Rényi efficiency, which penalizes distributions skewed toward frequent, low-content tokens and correlates more strongly with downstream results, though its generality remains debated~\citep{cognetta-etal-2024-two}. Other studies note that standard pre-tokenization pipelines often over-map pretokens to high-frequency words, reducing representational depth~\citep{reddy2025enough}.
Building on these efforts, \citet{bommarito2025kl3mtokenizersfamilydomainspecific} propose a framework encompassing token-per-character efficiency, domain coverage, and token size distribution. Collectively, these approaches point to a multidimensional view of tokenizer quality that integrates linguistic structure, statistical balance, and utility. \textbf{\textit{We distill this view into five core dimensions}} (see Appendix~\ref{app:core_evaluation} for examples). Each dimension emphasizes different evaluation levels: corpus-level diagnostics for fragmentation and alignment, and model-based probes for generalization and embedding use. Efficiency often trades off with alignment or interpretability, and priorities vary across applications. Developing benchmarks to quantify and balance these trade-offs remains an essential direction for future work.


\paragraph{Toward Practical Evaluation Methodologies.}
Translating these dimensions into practice requires consistent methodologies. Promising directions include unified corpus diagnostics, model-level robustness tests under distribution shift, and cross-lingual benchmarks linking tokenization quality to downstream performance. Standardized protocols would enable reproducibility and move the field beyond ad hoc, dataset-specific evaluation.
\begin{tcolorbox}[
colback=gray!7, colframe=black!11,
title=Key Evaluation Dimensions for Tokenizer Quality,
fonttitle=\bfseries\small, coltitle=black,
]
\vspace{-1mm} 
\small
\textbf{•}~\textbf{Coverage:} Does the tokenizer represent the linguistic units relevant to the target domain or language?

\textbf{•}~\textbf{Generalizability:} How well does it handle unseen forms, rare constructions, and morphologically complex variants?

\textbf{•}~\textbf{Linguistic Alignment:} Do token boundaries align with morphemes, affixes, or syntax?

\textbf{•}~\textbf{Robustness:} Is it resilient to variation, noise, and distributional shift (e.g., dialects, OCR errors)?

\textbf{•}~\textbf{Representation Utilization:} Are embeddings meaningfully activated during inference, or is much of the vocabulary underused?
\vspace{-1mm} 
\end{tcolorbox}

\paragraph{Open Trade-offs.}
As summarized in Table~\ref{tab:evaluation-standards}, evaluating tokenizers by only dev-set perplexity or vocabulary size obscures the real drivers of cost and quality. Report metrics that matter like tokens per 1k characters by language, parity ratios, compression–structure balance, and estimated pretrain FLOPs at fixed data budgets~\citep{goldman2024unpacking,ali-etal-2024-tokenizer,schmidt-etal-2024-tokenization}. Performance should be summarized across all data categories and shards used for training. Intrinsic metrics can triage candidates but need quick downstream probes to prevent gaming and preserve structural integrity~\citep{zouhar-etal-2023-tokenization,cognetta-etal-2024-two,uzan-etal-2024-greed}. Formal analyses and sizing analyses (explicit evaluations of how vocabulary size, sequence length, and token granularity affect memory use, throughput, and total training FLOPs) offer practical standards for selecting tokenizers that minimize total cost while maintaining target quality across languages~\citep{gowda-may-2020-finding,zouhar-etal-2023-formal,nayeem2025STRR}.



\if{false}
\paragraph{Call to Action: Building Shared Benchmarks}  
To advance tokenizer evaluation, we advocate for the development of shared, open-source benchmarks and diagnostic toolkits that:
\begin{itemize}
    \item Link tokenizer quality to downstream performance through task-based evaluation;
    \item Reflect linguistic and domain diversity, especially for underrepresented languages;
    \item Provide diagnostics for fragmentation, alignment, frequency balance, embedding activation, and stability across domains.
\end{itemize}

Such a framework will enable more principled, transparent, and reproducible evaluation—supporting better tokenizer design and adoption in both research and production settings.
\fi

\section{Conclusion}
The dominance of subword tokenization has fostered a one-size-fits-all mindset. Yet evidence from multilingual, low-resource, and domain-specific settings exposes its limits. Tokenization should be reimagined as a context-aware design choice, shaped by structure, data, task, and deployment constraints. This shift is vital for models that are efficient, scalable, fair, and linguistically grounded. Tokenization is not a preprocessing afterthought but a core component of model development, designed, evaluated, and refined alongside model behavior. 

\section*{Limitations}
Our argument centers on reframing tokenization as a modeling decision, but it remains primarily conceptual. We do not provide new empirical results or large-scale ablations, and thus our claims should be viewed as hypotheses for future validation and research. Implementing our proposed co-design and evaluation framework also assumes access to multilingual data, compute, and tooling that may not be universally available. Moreover, identifying standardized proxy probes that robustly predict large-scale downstream performance remains an open empirical problem, and we view this as a key direction for the future works. Finally, while we gesture toward token-free and byte-level approaches, these alternatives evolve too quickly for comprehensive inclusion here; our discussion focuses on guiding principles rather than definitive solutions. 

\section*{Acknowledgements}
We thank all the anonymous reviewers and the meta-reviewer for their valuable feedback and constructive suggestions for improving this work. Additionally, Mir Tafseer Nayeem acknowledges support from the Huawei PhD Fellowship, and Md Tahmid Rahman Laskar acknowledges the support from the CUPE 3903 Research Grant Fund, York University.

\bibliography{custom}

\clearpage
\appendix
\twocolumn[{%
 \centering
 \Large\bf Supplementary Material: Appendices \\ [20pt]
}]

\section{Illustrative Example for Token Fragmentation}
\label{illus:token_fragmentation}

\begin{figure*}[t]
    \centering
    \includegraphics[width=0.88\linewidth]{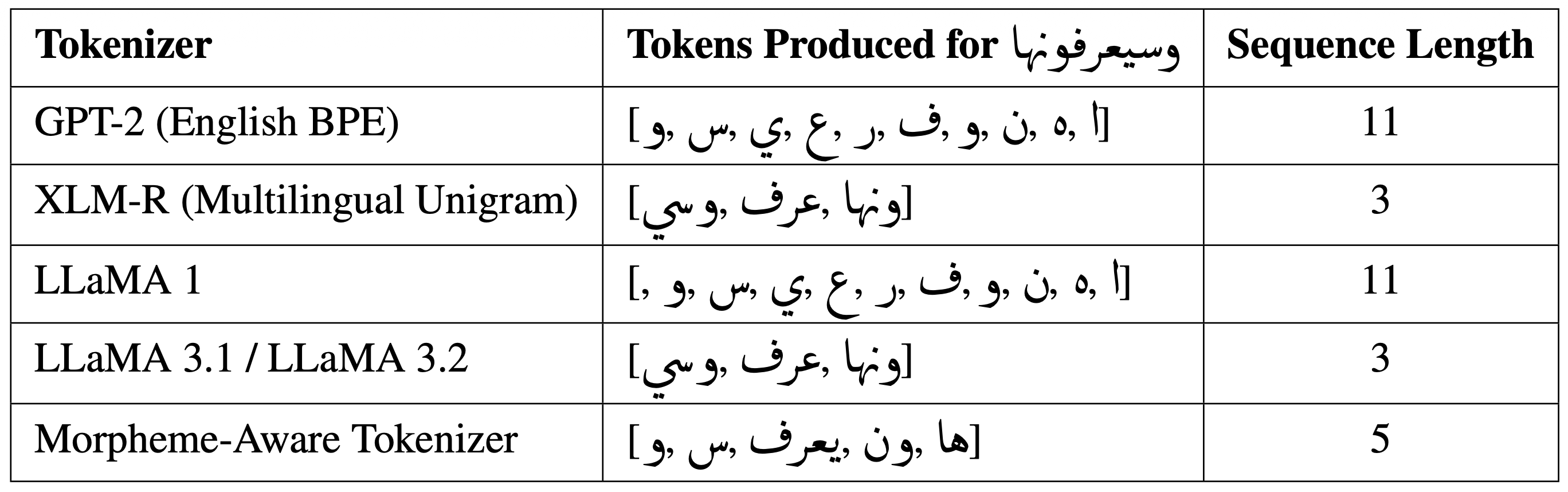}
    \caption{Tokenization of the Arabic word using different tokenization strategies.}
    \label{fig:arabic-tokenization}
\end{figure*}

A simple illustration of inefficient token fragmentation can be seen when applying a standard English-trained BPE tokenizer to morphologically rich languages such as Arabic. We emphasize that such tokenizers were not designed to support Arabic or other morphologically rich non-European languages, and are used here solely to illustrate the consequences of language–tokenizer misalignment rather than to imply a design failure. Consider the Arabic word in Figure~\ref{fig:arabic-tokenization} (``and they will recognize it''). This word combines a conjunction, a verb root, a subject marker, and an object pronoun, all within a single surface form. An English-centric tokenizer typically fails to capture this structure and instead fragments the word into arbitrary subword units.

In this example, the English-trained BPE tokenizer (GPT-2) produces eleven tokens, heavily fragmenting the Arabic word into individual characters and ignoring its internal morphological structure. The multilingual XLM-R tokenizer and the more recent LLaMA 3.1 model perform significantly better, segmenting the word into larger, semantically meaningful units. Interestingly, the morpheme-aware tokenizer results in more tokens than LLaMA 3.1, raising the question of whether fewer tokens always indicate better linguistic alignment or modeling efficiency. In contrast, the earlier LLaMA 1 tokenizer exhibits even greater over-fragmentation, treating spaces and letters as separate tokens and producing eleven fragments. Such over-fragmentation increases sequence length, reduces modeling efficiency, and disrupts semantic representation, effects that become especially pronounced in morphologically rich languages like Arabic, where single surface forms often encode multiple grammatical and semantic elements \citep{toraman2023impact,ali-etal-2024-tokenizer}. This over-fragmentation leads to higher tokenization costs and reduced model efficiency, particularly for non-English users who encounter faster context length limits and increased API costs.

\section{Evolution of Tokenization in Leading Models}
\label{appendix:evolution_tokenizer}
As discussed, the tokenizer is a fundamental component of model training, not merely an implementation detail. As LLMs evolve, future models must prioritize tokenization strategies to strengthen linguistic and logical capabilities. The following examples demonstrate that improving a language model often begins with revisiting and refining its vocabulary.

\subsection{GPT Series Tokenizer Advancements}

OpenAI's early GPT models, including GPT-2, used BPE tokenizer with a vocabulary of approximately 50k tokens \citep{radford2019language}. This tokenizer worked well for English but led to over-fragmentation of text in many other languages, causing significantly longer token sequences for non-English content (See   \S\ref{illus:token_fragmentation}). To address these issues, OpenAI improved the tokenizer in later models like ChatGPT (GPT-3.5) and GPT-4, introducing the cl100k\_base tokenizer with around 100k tokens, doubling the vocabulary size of GPT-2 \citep{Yang2024ProblematicTT}. This reduction in token count led to significant improvements. OpenAI continued refining the tokenizer with the 2024 "GPT-4o" update, which expanded the vocabulary to roughly 200k tokens. This change further reduced token usage, especially for non-English languages; however, this expansion came with a trade-off: the inclusion of rare tokens that were under-trained, leading to occasional issues like hallucinations or unexpected behavior, particularly with complex words in languages such as Chinese (See \S\ref{sec:responsible-tokenization}). 

\subsection{Tokenization Strategies in Google's Multilingual Models: PaLM, Gemma, and ByT5}
Google's PaLM (Pathways Language Model)~\citep{chowdhery2023palm} employs a SentencePiece tokenizer with a 256k vocabulary to handle a broad range of languages, reducing token fragmentation in languages like Vietnamese and Tamil. This strategy enhanced PaLM’s multilingual performance, particularly in tasks like translation. While the larger vocabulary increased memory usage, it provided improved language coverage. In 2024, Google released Gemma \citep{team2024gemma}, an open-access LLM with a redesigned tokenizer featuring a 262k vocabulary, also used in the Gemini models \citep{geminiteam2025geminifamilyhighlycapable}. This improved tokenizer mitigated over-tokenization and mis-splitting issues, enabling Gemma to manage over 100 languages and perform well in multilingual benchmarks, underscoring the key role of the tokenizer in its success, albeit with increased memory usage.

Google's ByT5 \citep{xue-etal-2022-byt5} takes a different approach by eliminating tokenization altogether, working directly on raw bytes to address challenges like out-of-vocabulary words, misspellings, and non-Latin scripts that subword tokenizers struggle with. ByT5 uses a 256-byte "vocabulary" (1 byte = 1 token), simplifying the process and ensuring comprehensive text coverage at the expense of longer sequences. While this increases sequence length, ByT5 reduces the model size for embeddings and improves robustness, particularly for noisy inputs. It outperformed tokenized models on tasks such as GLUE and QA, especially for smaller models, demonstrating the potential benefits of eliminating tokenization in specific scenarios. However, byte-level models, such as ByT5, face the trade-off of slower training and inference times due to longer sequence lengths. Despite this, Transformers remain computationally efficient, though ByT5 is slower during training and significantly slower during inference.

\subsection{LLaMA Series Tokenizer Advancements}
Meta’s LLaMA \citep{touvron2023llama} and LLaMA-2 \citep{touvron2023llama2} used a 32k BPE vocabulary primarily trained on English and European languages, leading to longer sequences, wasted context space, and poor performance on underrepresented languages (See \S\ref{illus:token_fragmentation}). Meta did not substantially redesign the tokenizer between LLaMA 1 and LLaMA 2, but improving multilingual performance in LLaMA 3 required changes beyond simple reuse, underscoring how tokenizer choices can necessitate retraining or substantial adaptation. With the release of LLaMA 3 and LLaMA 3.1 \citep{grattafiori2024llama3}, Meta addressed these issues by introducing a new tokenizer with a larger vocabulary and broader language coverage. The improved tokenizer reduces over-fragmentation in underrepresented languages and enhances performance across a wider range of languages. This update allows LLaMA 3 models to handle multilingual inputs more effectively, providing a more balanced representation of diverse languages and increasing model efficiency. However, these improvements come with the trade-off of higher memory usage and computation, which Meta optimized through architectural adjustments in LLaMA 3.1.

\section{Core Evaluation Dimensions}
\label{app:core_evaluation}

\textbf{(i) Coverage:} Evaluation must include coverage of rare words, domain-specific terminology, and named entities. Excessive fragmentation often indicates poor vocabulary alignment and leads to inefficiencies during modeling. For example, in the sentence "The patient was diagnosed with pheochromocytoma and referred to Dr. Al-Rahmani at the Hospital," a poorly designed tokenizer may over-segment the rare medical term pheochromocytoma or unnaturally split named entities like Al-Rahmani. High-coverage tokenizers should preserve such expressions with minimal degradation. As another example, for morphologically rich languages, tokenizer coverage can also be evaluated by comparing its output to that of a morphological analyzer. Measuring the divergence between tokenized segments and known morphemes can reveal how well the tokenizer captures meaningful subword units, contributing to both the depth (internal structure) and breadth (lexical variety) of vocabulary representation.

\vspace{2mm}

\noindent \textbf{(ii) Generalizability:} For instance, in the sentence ``She re-nationalized the industry after years of privatization,'' a well-generalizing tokenizer should segment re-nationalized in a way that reflects its compositional structure, even if the full form was absent from training. Tokenizers that treat related forms inconsistently can harm representation learning and impair downstream generalization.

\vspace{2mm}

\noindent \textbf{(iii) Linguistic Alignment:} Poor alignment obscures grammatical structure, reduces interpretability, and degrades performance in syntax-sensitive tasks. For example, Arabic verbs like yaktubūn (“they write”) and yarsimūn (“they draw”) share morphological patterns that should be preserved to support better generalization and cross-linguistic representation learning.

\vspace{2mm}

\noindent \textbf{(iv) Robustness:} Robust tokenizers should manage spelling errors, informal usage, dialectal forms, and code-switching without excessive fragmentation. In the mixed-language sentence "We discussed the results en la reunión this morning," a resilient tokenizer should treat the embedded Spanish phrase as a cohesive unit rather than fragmenting it unnaturally. Robustness is especially important in multilingual, real-world deployments~\cite{shohan-etal-2024-xl}.

\vspace{2mm}

\noindent \textbf{(v) Representation Utilization:} Evaluation must also consider which parts of the vocabulary and embedding space are actually used during inference. If a benchmark activates only 30\% of token embeddings, the majority of the learned vocabulary remains untested. Tokenizer evaluation should include coverage diagnostics that quantify how well a benchmark exercises the full representational capacity of the model.

\section{Usage of Large Language Models}
\label{sec:LLM-usage}
We disclose that large language models were used in limited, assistive roles. Specifically, they supported \textbf{text polishing}, including improvements to grammar, spelling, phrasing, and word choice, with all suggestions reviewed by the authors. All outputs were manually verified, and the authors remain fully responsible for the research content and conclusions.

\begin{table*}[t]
\centering
\small
\setlength{\tabcolsep}{3pt}
\renewcommand{\arraystretch}{1.12}
\begin{tabularx}{\textwidth}{PYYYY}
\toprule
\textbf{Paper (year)} & \textbf{Efficiency} & \textbf{Linguistic fidelity} & \textbf{Fairness} & \textbf{Computational cost} \\ \hline
\midrule
\citet{sennrich-etal-2016-neural} &
High compression compared to characters, shorter sequences than characters or bytes. &
Greedy merges can split morphemes, alignment varies by language. &
English-centric corpora bias split patterns across languages. &
Fewer tokens lower attention cost, tokenizer training is offline, embedding size grows with vocabulary. \\ \hline
\citet{kudo-2018-subword} &
Compression similar to BPE with improved robustness from sampling. &
Unigram often respects morphemes better than greedy BPE. &
Less brittle across scripts than word level. &
Training adds sampling on the fly, inference similar to BPE, FLOPs comparable at a fixed sequence length. \\ \hline
\citet{kudo-richardson-2018-sentencepiece} &
Fast raw-text training and deployment. &
Supports BPE or Unigram with normalization that affects morphology. &
Behavior depends on corpus balance and normalization choices. &
Minimal runtime overhead, compute impact is via vocabulary size and sequence length. \\ \hline
\citet{tay2022charformer} &
Learns blocks from characters, downsampling yields near-subword compression. &
Improved compositionality from learned subwords over raw chars. &
More uniform across scripts than fixed subword vocabularies. &
Light module overhead but shorter effective sequences reduce FLOPs versus char or byte inputs. \\ \hline
\citet{libovicky-helcl-2024-lexically} &
Uses morphological analyzers to guide merges at training time. &
Better morpheme alignment than frequency-only segmentation. &
Benefits morph-rich and low-resource languages. &
Heavier tokenizer training, similar inference token counts to BPE, can reduce wasted tokens where over-segmentation existed. \\ \hline
\citet{chizhov-etal-2024-bpe} &
Prunes low-value merges to improve compression and utility. &
Cleaner boundaries and fewer under-trained tokens. &
Reduces fragmentation disparities in multilingual settings. &
Small extra tokenizer-training cost, shorter sequences give modest FLOP savings. \\ \hline
\citet{zouhar-etal-2023-formal} &
Faster implementations and complexity insights. &
Theoretical guidance for better merge schedules. &
\textbf{\textit{Not Available}}  &
Lower tokenizer build cost, model compute unchanged except via sequence length. \\ \hline
\citet{gowda-may-2020-finding} &
Identifies a sweet spot balancing compression and generalization. &
Very small vocabularies harm morphology, very large vocabularies memorize idiosyncrasies. &
Dominant-language tuning can harm others. &
Too small vocabularies increase sequence length, too large increase embedding parameters, choose V to minimize total compute. \\ \hline
\citet{asgari2025morphbpe} &
Slightly less compression than aggressive BPE with better utility. &
Blocks merges across morpheme boundaries for higher consistency. &
Improves parity for morph-rich languages by reducing fragmentation. &
Minimal integration cost, sequence length can rise slightly but faster convergence often offsets compute. \\ 
\bottomrule
\end{tabularx}
\caption{Training Tokenizers from Scratch: core algorithms, learned tokenizers, and theory that shape efficiency, fidelity, and compute from the ground up.}
\label{tab:train-from-scratch}
\end{table*}

\begin{table*}[t]
\centering
\small
\setlength{\tabcolsep}{3pt}
\renewcommand{\arraystretch}{1.12}
\begin{tabularx}{\textwidth}{PYYYY}
\toprule
\textbf{Paper (year)} & \textbf{Efficiency} & \textbf{Linguistic fidelity} & \textbf{Fairness} & \textbf{Computational cost} \\ \hline
\midrule
\citet{pfeiffer-etal-2021-unks} &
Reduces UNK rates on unseen scripts, dedicated in-language tokenizer yields shorter sequences than character fallback. &
Captures script-specific orthography, aligned embeddings preserve word-level semantics. &
Targets under-served languages with unseen scripts, narrows cross-language token-length disparities. &
Offline tokenizer build plus small added embeddings, no full model retraining, inference cost similar while fewer tokens reduce attention and cache compute. \\ \hline
\citet{rust-etal-2021-good} &
Monolingual tokenizers improve efficiency for target languages. &
Better alignment and accuracy for the intended language. &
Multilingual vocabularies show monolingual bias and parity issues. &
Retokenizing and re-embedding have low overhead compared to pretraining, shorter sequences reduce FLOPs. \\ \hline
\citet{ali-etal-2024-tokenizer} &
Tokenizer choice materially shifts compression and utility. &
Fertility and parity alone are weak quality predictors. &
English-centric designs raise costs for many languages. &
Inefficient vocabularies lengthen sequences and steps, better multilingual tokenizers lower training and inference compute. \\ \hline
\citet{yehezkel-pinter-2023-incorporating} &
Slightly less compression than pure BPE with better utility. &
Context-aware subwords improve cohesion and morphology. &
Reduces over-segmentation in agglutinative languages. &
Heavier offline build only, runtime similar to subword, potential token reductions yield modest savings. \\ \hline
\citet{chizhov-etal-2024-bpe} &
Merge pruning improves compression and quality. &
Cleaner boundaries and fewer under-trained tokens. &
Less fragmentation across languages. &
Small tokenizer-training overhead, shorter sequences save compute. \\ \hline
\citet{nguyen2023seallms} &
Vocabulary expansion merges NLLB tokens into LLaMA for SEA languages. &
Fewer unnatural splits in non-Latin scripts with better word coverage. &
Directly reduces cross-language token length disparities while preserving English efficiency. &
Expanded vocabularies increase embedding size, fewer tokens cut attention and cache cost, later versions reuse large Gemma vocabularies. \\ \hline
\citet{bari2025allam} &
Tokenizer augmentation improves Arabic token economy. &
Arabic-aware subwords better match morphemes while preserving English skill. &
Narrows parity gaps for Arabic versus English. &
Embedding and softmax grow with expansion, fewer Arabic tokens reduce FLOPs per step, tokenizer work is offline. \\ \hline
\citet{team2025fanar} &
From-scratch and continued-pretrain models show practical adaptation. &
Arabic-centric training improves segmentation for Arabic forms. &
Improves parity for Arabic in a bilingual setup. &
Reusing Gemma tokenizer avoids costly remapping, larger vocab raises memory, fewer tokens reduce attention cost. \\ \hline
\citet{uzan-etal-2024-greed} &
Intrinsic measures reduce expensive full model comparisons. &
Segmentation quality can be assessed without retraining models. &
Better selection reduces bias from ad-hoc choices. &
Cuts search compute by avoiding multiple retrains during adaptation. \\ \hline
\citet{schmidt-etal-2024-tokenization} &
Compression alone is not a sufficient design signal. &
Emphasizes morphology and semantics beyond length. &
Warns that compression-driven reuse can harm parity. &
Guidance avoids wasted pretraining and encourages retokenization where it reduces steps. \\ 
\bottomrule
\end{tabularx}
\caption{Adapting Tokenizers: rethinking reuse through language expansion, refinement, and context-aware construction to cut tokens and cost in new domains.}
\label{tab:adapt-reuse}
\end{table*}

\begin{table*}[t]
\centering
\small
\setlength{\tabcolsep}{3pt}
\renewcommand{\arraystretch}{1.12}
\begin{tabularx}{\textwidth}{PYYYY}
\toprule
\textbf{Paper (year)} & \textbf{Efficiency} & \textbf{Linguistic fidelity} & \textbf{Fairness} & \textbf{Computational cost} \\ \hline
\midrule
\citet{petrov2023language} &
Documents large cross-language token count gaps for the same content. &
Splits often ignore morphology outside Latin scripts. &
Up to large disparities with pricing and latency inequities. &
Over-tokenized languages pay higher FLOPs per word and lose effective context, fairer tokenizers reduce compute. \\ \hline
\citet{ahia-etal-2023-languages} &
API cost varies with tokenizer-induced length differences. &
Over-fragmentation lowers utility for some languages. &
Identifies systematic overcharging of certain scripts. &
Parity-aware tokenizers reduce steps and tokens, inference cost scales with tokens so parity improves efficiency. \\ \hline
\citet{chai-etal-2024-tokenization} &
Segmentation is fragile under typos and format shifts. &
LLMs are sensitive to token boundaries, dropout improves robustness. &
Variants can disproportionately affect morph-rich languages. &
Robustness training adds small CPU overhead, inference unchanged, avoids token length blowups in the wild. \\ \hline
\citet{asgari2025morphbpe} &
Competitive compression with better utility in difficult scripts. &
Prevents merges across morpheme boundaries. &
Improves parity for morph-rich languages by reducing fragmentation. &
Slight sequence length increase can be offset by faster convergence, overall compute does not rise materially. \\ \hline
\citet{chizhov-etal-2024-bpe} &
Pruning merges improves compression and quality. &
Cleaner boundaries reduce pathological splits. &
Less fragmentation in multilingual settings. &
Token counts drop slightly, giving modest FLOP savings. \\ \hline
\citet{nguyen2023seallms} &
Vocabulary expansion improves token economy in SEA languages. &
Fewer unnatural splits for non-Latin scripts. &
Shrinks token length disparities while preserving English performance. &
Embedding growth trades for fewer tokens, net reduction in attention compute for SEA text. \\ \hline
\citet{bari2025allam} &
Arabic tokenization improved through expansion and augmentation. &
Arabic-aware units better match morphology. &
Narrowed parity gap for Arabic. &
Fewer Arabic tokens reduce step compute, expansion cost is offline. \\ \hline
\citet{rust-etal-2021-good} &
Monolingual tokenizers are more efficient for specific languages. &
Better alignment for the target language. &
Multilingual vocabularies show parity issues. &
Retokenization overhead is small compared to the compute saved by shorter sequences. \\
\bottomrule
\end{tabularx}
\caption{Responsible tokenization addressing parity, robustness, and stability to reduce unequal costs and failures across languages and noisy inputs.}
\label{tab:responsible}
\end{table*}

\begin{table*}[t]
\centering
\small
\setlength{\tabcolsep}{3pt}
\renewcommand{\arraystretch}{1.12}
\begin{tabularx}{\textwidth}{PYYYY}
\toprule
\textbf{Paper (year)} & \textbf{Efficiency} & \textbf{Linguistic fidelity} & \textbf{Fairness} & \textbf{Computational cost} \\ \hline
\midrule
\citet{goldman2024unpacking} &
Higher compression often correlates with better generation quality at smaller scales. &
Compression is helpful but structure also matters. &
Over-compressed languages can lose linguistic structure. &
Fewer tokens reduce FLOPs, gains saturate at large scale, report both compression and structure. \\ \hline
\citet{zouhar-etal-2023-tokenization} &
Intrinsic metric correlates with downstream in some settings. &
Penalizes overly common or overly rare units to encourage balance. &
Can guide parity-aware design if used cautiously. &
No model-time cost, helps pick vocabularies that reduce sequence length or maintain quality at same compute. \\ \hline
\citet{cognetta-etal-2024-two} &
Shows intrinsic metrics can be gamed without real gains. &
High Rényi scores do not always imply better segmentation. &
Warns against proxy metrics. &
Avoids wasted pretraining on misleading metrics, no runtime change. \\ \hline
\citet{uzan-etal-2024-greed} &
Compares intrinsic measures to avoid repeated full retrains. &
Assesses segmentation quality without a model in the loop. &
Selection procedures can reduce bias from ad-hoc choices. &
Cuts search compute during tokenizer selection, enables faster iteration. \\ \hline
\citet{schmidt-etal-2024-tokenization} &
Compression alone is not a sufficient signal for design. &
Emphasizes morphology and semantics in evaluation. &
Warns that compression-driven design can harm parity. &
Guidance reduces wasted compute and improves selection efficiency. \\ \hline
\citet{zouhar-etal-2023-formal} &
Complexity and implementation analysis for BPE. &
Theoretical guidance for constraints and schedules. &
\textbf{\textit{Not Available}} &
Tokenizer build cost reduced, model compute mediated by sequence length only. \\ \hline
\citet{gowda-may-2020-finding} &
Locates vocabulary size that balances compression and generalization. &
Very small or very large vocabularies degrade linguistic structure. &
\textbf{\textit{Not Available}} &
Minimizes total compute by trading embedding size against sequence length, report chosen V with rationale. \\ \hline
\citet{ali-etal-2024-tokenizer} &
Shows tokenizer choice materially affects cost and quality. &
Finds weak correlation of some popular intrinsic indicators with final quality. &
Highlights cross-language disparities from English-centric design. &
Encourages reporting token counts, length parity, and training cost impacts alongside quality. \\ 
\bottomrule
\end{tabularx}
\caption{Evaluation Framework and Missing Standards of metrics, analyses, and protocols that make tokenizer selection measurable, comparable, and compute-aware.}
\label{tab:evaluation-standards}
\end{table*}

\begin{table*}[t]
\centering
\small
\begin{tabular}{p{0.26\linewidth} p{0.70\linewidth}}
\toprule
\textbf{Stage} & \textbf{Guiding Questions} \\ \hline
\midrule
\textbf{Training from Scratch} &
(i) Do target languages over-fragment under English-centric vocabularies? \newline
(ii) Which vocabulary size minimizes end-to-end FLOPs rather than dev-set perplexity? \newline
(iii) Can morphology-aware constraints reduce token counts in dominant languages? \newline
(iv) Could a character/byte backbone with learned downsampling add robustness without higher compute cost? \\[0.5em]
\hline

\textbf{Adapting Existing Tokenizers} &
(i) Where is token inflation worst, and what parity ratio is acceptable in production? \newline
(ii) Is a small vocabulary expansion sufficient, or is a dedicated tokenizer needed? \newline
(iii) Can checkpoint compatibility be preserved when modifying the tokenizer (e.g., mapped or appended embeddings)? \newline
(iv) What intrinsic or downstream probes will you run before a full retrain? \newline
(v) How will you attribute compute savings to shorter sequences versus parameter changes? \\

\hline
\textbf{Responsible Tokenization} & 
(i) What token-length disparity across top languages is acceptable, and how will you enforce it? \newline
(ii) Do evaluations include noisy, code-switched, or adversarial text to test segmentation stability? \newline
(iii) Which mitigation (dropout, merge refinement, morphology-aware rules, or language-specific expansion) yields the largest token reduction for the weakest languages? \newline
(iv) How will you track pricing and latency equity linked to tokenizer-induced length differences? \newline
(v) What rollout and monitoring plan will detect regressions in parity or robustness after deployment? \\
\hline
\textbf{Evaluation Framework and Missing Standards} & 
(i) Which metrics act as pass/fail gates for selection—tokens per 1k characters, parity ratios, compression–structure measures, or estimated FLOPs deltas? \newline 
(ii) What small, repeatable downstream tasks will validate candidates beyond intrinsic metrics? \newline 
(iii) How will you disentangle and report the effects of sequence length versus embedding size on step time and memory? \newline 
(iv) What decision rule determines when to stop iterating and lock the tokenizer for production? \newline 
(v) How will you document tokenizer choices to ensure cost and quality claims are reproducible? \\
\bottomrule
\end{tabular}
\caption{Consolidated guiding questions for principled tokenizer development and adaptation. Each stage highlights key diagnostic prompts for evaluating design trade-offs, resource efficiency, and cross-lingual robustness.}
\label{tab:guiding_questions}
\end{table*}

\end{document}